\documentclass{article}

\usepackage{microtype}
\usepackage{graphicx}
\usepackage{subfigure}
\usepackage{booktabs} 

\usepackage{hyperref}

\usepackage[accepted]{icml2018}

\icmltitlerunning{Adversarial Active Learning}

\newcommand{\etal}{\textit{et al}.~}

\newcommand{\mnist}{\textit{MNIST}~}
\newcommand{\shoes}{\textit{Shoe-Bag}~}
\newcommand{\quick}{\textit{Quick-Draw}~}

\newcommand{\dfal}{DFAL~}
\newcommand{\bald}{BALD~}
\newcommand{\egl}{EGL~}
\newcommand{\ceal}{CEAL~}
\newcommand{\core}{CORE-SET~}
\newcommand{\uncertainty}{uncertainty~}
\newcommand{\random}{RANDOM~}

\newcommand{\lenet}{LeNet5~}
\newcommand{\vgg}{VGG8~}

\begin{document}

\twocolumn[
\icmltitle{Adversarial Active Learning for Deep Networks: \\
           a Margin Based Approach}

\begin{icmlauthorlist}
\icmlauthor{Melanie Ducoffe}{to}
\icmlauthor{Frederic Precioso}{to}
\end{icmlauthorlist}

\icmlaffiliation{to}{University of Nice Cote d\'Azur, - I3S, UMR UNS-CNRS 7271 06900 Sophia Antipolis, France}

\icmlcorrespondingauthor{Melanie Ducoffe}{ducoffe@i3s.unice.fr}

\icmlkeywords{Adversarial Attacks, Active Learning, Deep Learning, Margin}

\vskip 0.3in
]



\printAffiliationsAndNotice{\icmlEqualContribution} 
\begin{abstract}
We propose a new active learning strategy designed for deep neural networks.
The goal is to minimize the number of data annotation queried from an oracle during training.
Previous active learning strategies scalable for deep networks were mostly based on uncertain sample selection.
In this work, we focus on examples lying close to the decision boundary.
Based on theoretical works on margin theory for active learning, we know that such examples may help to considerably decrease the number of annotations.
While measuring the exact distance to the decision boundaries is intractable, we propose to rely on adversarial examples. We do not consider anymore them as a threat instead we exploit the information they provide on the distribution of the input space in order to approximate the distance to decision boundaries.
We demonstrate empirically that adversarial active queries yield faster convergence of CNNs trained on \textit{MNIST}, the \textit{Shoe-Bag} and the \textit{Quick-Draw} datasets.
\end{abstract}
\section{Introduction}
The efficiency of deep networks is mainly known under typical training procedures and with large datasets. 
However, gathering and annotating huge dataset for supervised learning may prohibit the expansion of deep networks 
towards new fields such as chemistry or medicine \cite{2018arXiv180109319S,hoi2006batch}. A possible solution to build online an efficient but reduced training set is to rely on active learning.
Active learning is a family of methods seeking to optimize automatically the training set for the task at hand in order to limit the need of human annotation. 
Active learning strategies are not only motivated by theoretical works demonstrating that one model may perform better using less labeled data if the data are model-crafted \cite{cohn1996active}, but also by its proven efficiency on a wide range of machine learning procedures: from preference rating information for a new user 
in a movie recommendation system \cite{sun2013learning} to classifying medical data that often requires very high cost labeling \cite{hoi2006batch}.
It is only recently that active learning has been investigated on deep networks, especially CNNs. The question to scale active learning on deep networks has been raised on a diverse range of topics: from image classification, to sentiment classification, 
or to VQA and dialogue generation \cite{Gal2016Active, zhou2010active, 2017arXiv171101732L, asghar2017deep}. All those works converge to a common assessment on the efficiency of active 
learning to reduce the need for a large labeled training set.
Yet, transposing directly existing active learning on deep networks is not intuitive. First of all, scaling them to high dimensional parameters 
networks may turn out to be intractable: some classic active learning methods such as Optimal Experiment Design \cite{yu2006active} require to inverse the Hessian matrix of the models at each iteration, which would be intractable for current standard CNNs.
Secondly, one of the most standard active learning strategy is to rely on uncertainty measure. Uncertainty in deep networks is usually evaluated through the network's output however this is known to be misleading. Indeed, the discovery of adversarial examples has demonstrated that the way we are measuring uncertainty may be overconfident. 
Adversarial examples are inputs modified with small (sometimes not perceptually distinguishable) but specific perturbations which result 
in an unexpected misclassification despite a strong confidence of the network in the predicted class\cite{szegedy2013intriguing}. On one hand, the existence of such adversarial examples somehow discards uncertainty-based selection from being an efficient active learning criterion for deep networks.
On the other hand, the magnitude of adversarial attacks does provide an information about how far a sample is from the decision boundaries of a deep network. 
This information is relevant in active learning and known as margin-based active learning. In a generic margin-based active learning, we assume that the decision boundaries evolve towards the optimal solution as the training set increases. Hence samples lying the farthest from the decision boundaries do not need to be labeled by a human expert, as long as the current model is consistent in its predictions with the optimal solution. In order to refine the current model, margin-based active learning queries the unlabeled samples lying close to the decision boundary. 
Balcan \etal, in \cite{balcan2007margin}, has demonstrated the significant benefit of margin-based approaches in reducing human annotations: in specific cases, one 
may obtain an exponential improvement over human labeling. However, it requires 
computing the distance between a sample and the decision boundaries which is not tractable
when considering deep networks. Although we can approximate this distance by considering the minimal distance between two samples from different classification regions (i.e. corresponding to two different classes), such an evaluation is computationally expensive, nor it provides a close upper bound to the real criterion. Eventually, the minimal adversarial perturbation of a sample 
does provide a better upper bound on how far this sample is from the decision boundaries.

In this article, we do not consider adversarial examples as a threat but rather as a guidance tool to query new data. 
Our work focuses on a new active selection criterion based on the sensitiveness of unlabeled examples to adversarial attacks.
Specifically, our contributions are twofold:

\textbullet~We present a new heuristic for margin-based active learning for deep networks, called DeepFool Active Learning method (\dfal).
It queries the unlabeled samples, which are the closest to their adversarial attacks, labels not only the unlabeled sample but its adversarial counterparts as well, using twice the same label. This pseudo-labeling comes for free without introducing any corrupted labels in the training set.\\
\textbullet~We empirically demonstrate that \dfal labeled data may be used on other networks than the one they have been designed for, while achieving higher accuracy than random selection. To the best of our knowledge, this is the first active learning method for deep networks tested for this property.

We describe other active learning methods in the section \textit{Related work}. The following section, \textit{Adversarial Active Learning 
with Deep-Fool attacks}, describes our method \dfal. Finally, in \textit{Experiments}, we demonstrate empirically
the efficiency of our algorithm on three datasets that have been considered in recent methods on active learning for deep networks: \mnist, \quick, and \shoes. Not only we achieve state-of-the-art accuracy on those three tasks, 
but our methods run much faster than the previous state-of-the-art approaches.

\section{Related Work}

For a review of classic active learning methods and their applications, we refer the reader to Burr Settles \cite{settles2010active}.
The main principle of active learning methods lies in iteratively building the training set: the iterative process alternates between training the classifier on the current labeled training set, and after convergence of the model, asks an oracle (usually a human annotator) 
to label a new set of points. Those new points are queried from a pool of unlabeled data given the heuristic in use. 
Several heuristics coexist as it is impossible to obtain a universal active learning strategy effective for any given task \cite{dasgupta2005analysis}. 
When it comes to deep learning, especially CNN, many existing active learning heuristics have proven to be not effective.
For example, we empirically noticed in our experiments that uncertainty selection, or uncertainty sampling \cite{lewis1994sequential}, may perform worse than passive random selection. Since uncertainty selection consists in querying the annotations for the unlabeled samples which lead to predictions with lowest confidence, its cost is low and its setup simple. It has thus been used on deep networks for various tasks, ranging from sentiment classification to visual question answering and Named Entity Recognition \cite{zhou2010active, 2017arXiv171101732L, shen2018deep}. 
Uncertainty selection has been improved in a pseudo-labeling method called \ceal \cite{wang2016cost}: \ceal performs uncertainty selection, but also adds highly confident samples into the increased training set. The labels of these samples are not queried but infered from the network's predictions. In the case, one deal with a highly accurate network, CEAL will definitely improve the generalization accuracy. However, CEAL implies new hyperparameters to threshold the prediction's confidence. If such a threshold is badly tuned, 
it will corrupt the training set with mistaken labels.
Uncertainty selection may be also tailored to network ensemble, either by disagreement over the models (\textit{Query by committee}, \cite{seung1992query}) 
or by sampling through the distribution of the weights (\textit{Bayesian active learning}, \cite{kapoor2007active}). Recently, Gal \etal, in \cite{Gal2016Active},  demonstrated that dropout
(and other stochastic regularization schemes) is equivalent to perform inference on the posterior distribution of the weights, enabling to leverage the cost of training and updating multiple models. 
Thus, dropout allows to sample an ensemble of models at test time: to perform \textit{Dropout Query By Committee} (Ducoffe \etal, \cite{ducoffe2015qbdc}) or \textit{Bayesian Active Learning} (Gal \etal, \cite{Gal2016Active}). 
Gal \etal proceeded with a comparison of several active learning heuristics: among all the metrics, \bald which maximizes the mutual information between predictions and model posterior consistently outperforms other metrics.

In the original problem, active learning only queries one sample at a time. However, such a strategy would not be stable considering deep networks. 
Since CNNs, and other deep learning algorithms, are trained with local optimization schemes, we need to add several sample at a time to have a consistent 
impact on the training. A possible solution is to select the samples with the top scores.

Sener \etal \cite{sener2018active} define the batch active learning problem as a core set selection. They minimize the population risk of a model learned on a small labeled subset. To do so they propose an upper bound with a linear combination 
of the training error, the generalization error and a third term denoted as the core set loss. Due to the expressive power of CNNs, 
the authors argue that the first two terms (training and generalization error) are negligible. Therefore the population risk would mainly be controlled by the core set loss. 
The core set loss consists in the difference between the average empirical loss over the set of points which are already labeled, and the average empirical loss over the entire dataset including unlabeled points. 
If not considering the labels, the core set loss is equivalent to computing the covering radius over the network prediction. Finally, Sener \etal used a mixed integer programming heuristic to minimize at best the covering radius of the data. Thanks to their method, they achieve state-of-the-art performance in active learning for image classification.

Another direction, rarely explored for deep networks, is to rely on the distance to decision boundaries, 
namely margin-based active learning. Assuming that the problem is separable with a margin is a reasonable requirement assumed for many popular models such as SVM, Perceptron or AdaBoost.
When positive and negative data are separable under SVM, Tong \etal have demonstrated the efficiency of picking the example which is the closest to the decision boundary \cite{tong2001support}.
If, exploiting the geometric distances has been relevant for active learning on SVM \cite{tong2001support,brinker2003incorporating}, 
it is not intuitive for CNNs since we do not know beforehand the geometrical shape of their decision boundaries. A first trial has been proposed in \cite{Zhang2017}. 
The Expected-Gradient-Length strategy (\egl) consists in selecting instances with a high magnitude gradient. 
Not only such samples will have an impact on the current model parameter estimates but they will likely modify the shape of the decision boundaries. However, 
computing the true gradient for a given sample is intractable without its ground-truth label. 
In practice, they approximate the gradient with the expectation over the gradients conditioned on every possible class assignments.
\section{Adversarial Active Learning with Deep-Fool attacks}

\begin{figure}[!h]
\centering
 \subfigure[Shortest distance to the boundary]{
  \includegraphics[width=0.22\textwidth, height=0.15\textheight]{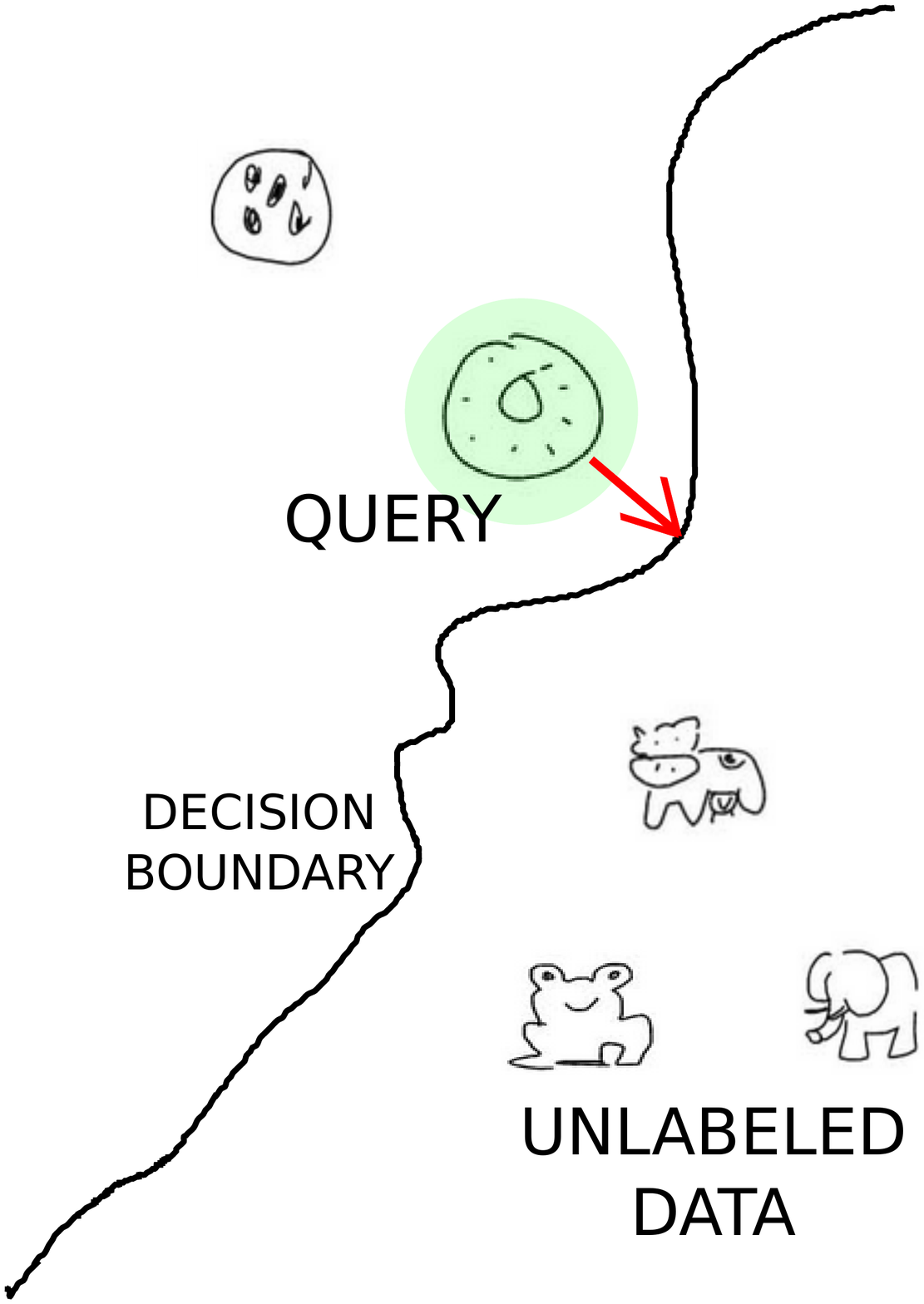}
  \label{fig:schema_0}
 }
 \subfigure[Approximation by the distance to the closest any-other-class sample]{
  \includegraphics[width=0.22\textwidth, height=0.15\textheight]{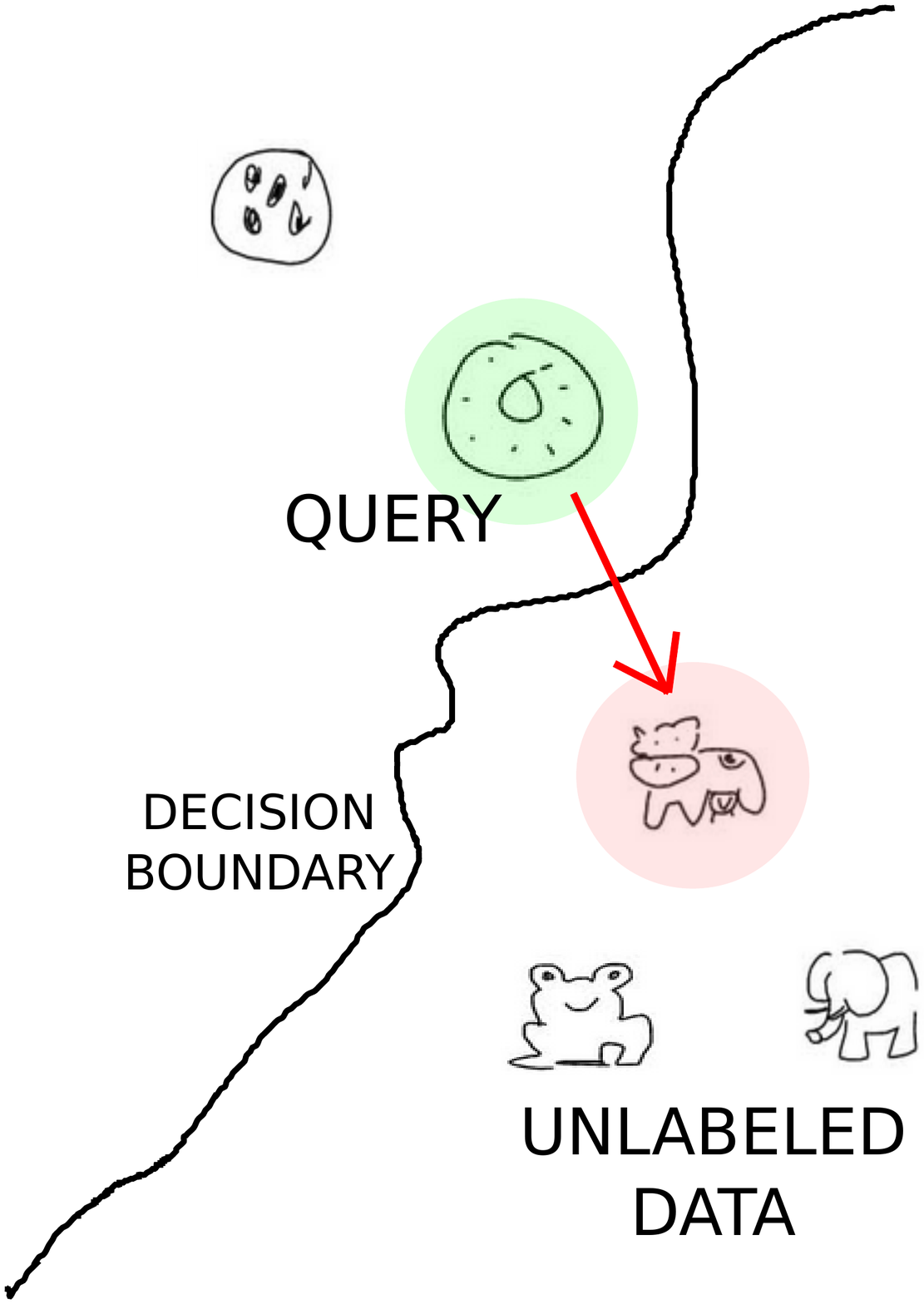}
  \label{fig:schema_1}
 }
 
 \subfigure[Approximation by the distance to the adversarial example]{
  \includegraphics[width=0.22\textwidth, height=0.15\textheight]{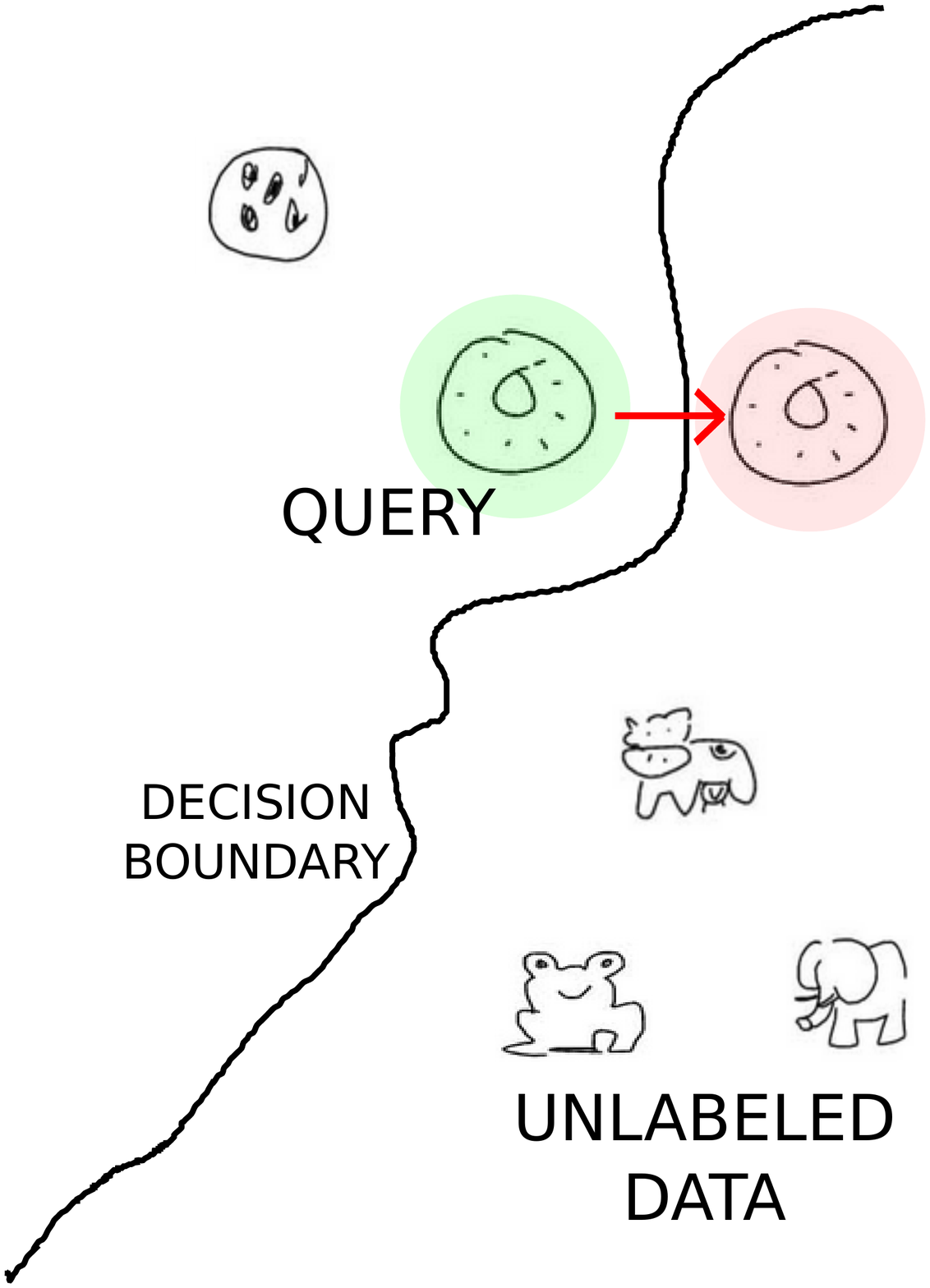}
  \label{fig:schema_2}
 }
 \subfigure[DFAL Strategy]{
  \includegraphics[width=0.22\textwidth, height=0.15\textheight]{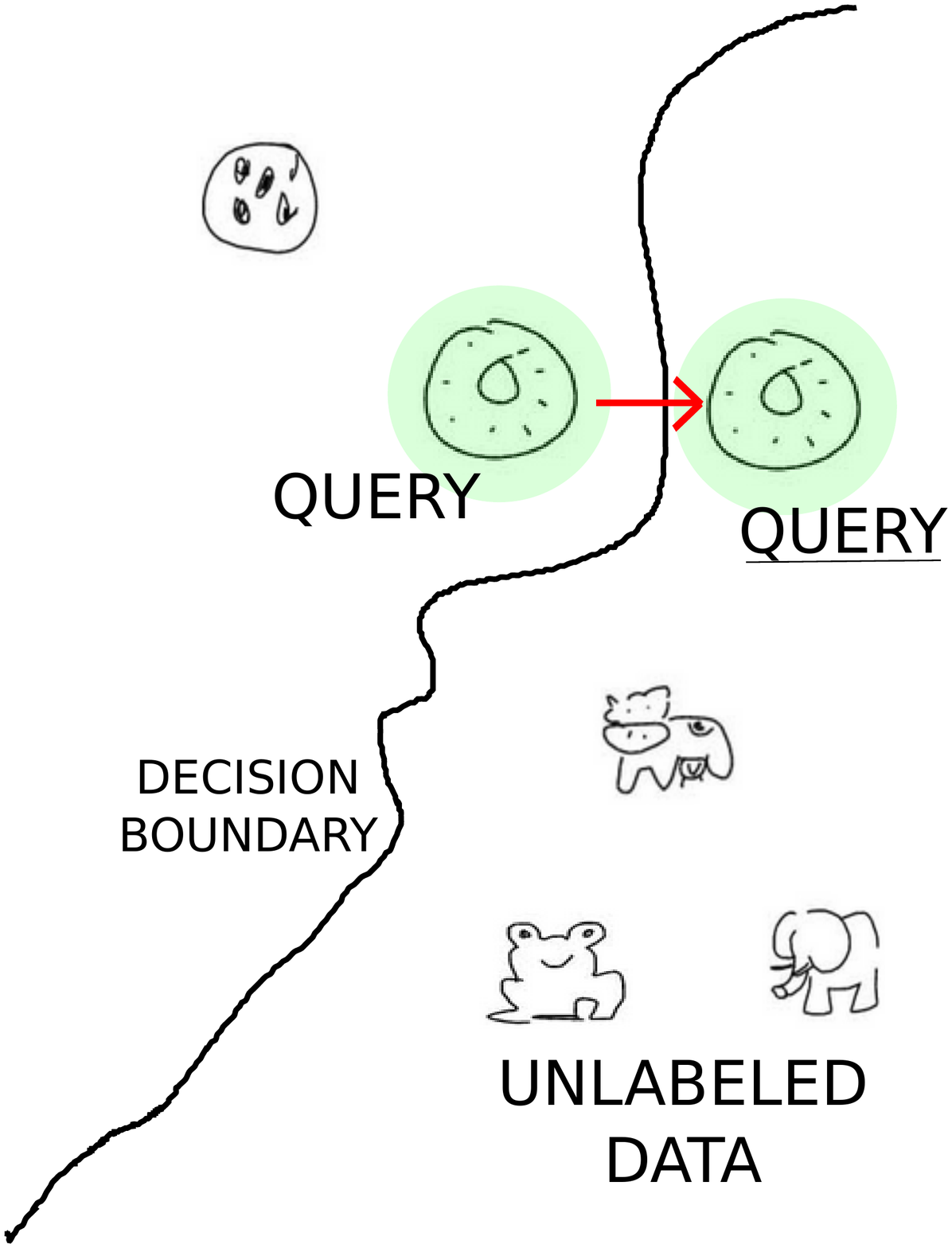}
  \label{fig:schema_3}
  }
 \caption{Illustration of different margin-based active learning scenarios in the binary case}
 \label{fig:schema}
\end{figure}

In \cite{balcan2007margin}, Balcan \etal demonstrated the significant benefit of margin-based approaches in reducing human annotations.
We illustrate several margin-based active learning heuristics in figure~\ref{fig:schema}: for each scenario, the data underlined in green will be queried. Especially, figure~\ref{fig:schema_3} describes our contribution.
In the original case in figure~\ref{fig:schema_0}, the projection of an unlabeled sample to the decision boundary determines whether or not it is worth to query its label, depending on the distance between the sample and the boundary.
Margin-based strategies are effective but they require to know how to compute the distance to the decision boundary. When such a distance is intractable, a naive approximation  consists in computing instead the distance between the sample of interest and its closest neighboring sample which has a different predicted class.
\vspace{-0.5cm}

Approximating the distance between a sample and the decision boundary, by the distance between this same sample and its closest neighboring sample from a different class, is coarse and computationally expensive.\\
Instead, we propose \dfal, a Deep-Fool based Active Learning strategy which selects unlabeled samples with the smallest adversarial perturbation.

%
%
Indeed, adversarial attacks were originally designed to approximate the smallest perturbation to cross the decision boundary. Hence, in a binary case, the
distance between a sample and its smallest adversarial example better approximates the original distance to the decision boundary than the aforementioned approximation, as illustrated in figure~\ref{fig:schema_2}. In a binary case, the label of the sample added to the training set is then given by the network prediction. Usually, adversarial attacks which would allow us to design a perturbation requires also to know the target label however in a binary case the target class of the attack is obvious.

In a multi-class context everything is different: we do not have any prior knowledge on which class the closest adversarial region belongs to. 
Inspired from the strategy done previously in EGL~\cite{Zhang2017}, we could design as many perturbations as the number of classes and keep only the smallest perturbation, but this would be time consuming. The EGL approach is then discarded.\\
We thus have to consider the available techniques of adversarial attacks from the literature \cite{szegedy2013intriguing,Goodfellow2015,carlini2016defensive} and look for the hardest technique to counter since it will provide more information on the margin in more cases and in more difficult cases. To the best of our knowledge, Carlini \etal~\cite{carlini2017towards, 206180, Carlini2017AEE} methods are among the hardest attacks to counter. However, it also requires to tune several hyperparameters.

We have thus decided to use \textit{Deep-Fool} algorithm to compute adversarial attacks for \dfal \cite{moosavi2016deepfool}. Indeed, \textit{Deep-Fool} is an iterative procedure which alternates between a local linear approximation of the classifier around the source sample and an update of this sample so that it crosses the local linear decision. 
The algorithm stops when the updated source sample becomes effectively an adversarial sample regarding the initial class of the source sample.
When it comes to \dfal, \textit{Deep-Fool} holds three main advantages: (i) it is hyperparameter free (especially it does not need target labels which makes it more compliant with multi-class contexts); (ii) it runs fast as we empirically noticed in table~\ref{tab:complexity}; (iii) it is competitive with state-of-the-art adversarial attacks.

Moreover, \dfal is theoretically motivated by the robustness of neural networks: in \cite{zahavy2018ensemble}, Xu \etal used robustness to explain the generalization abilities of stochastic algorithms. They can generalize well as long as their sensitiveness to adversarial examples is bounded in average. Xu \etal explain that since deep learning methods, in the majority of cases, are involving stochastic optimization mechanisms due to the common schemes used in their training phase such as SGD or dropout, they can be considered as stochastic algorithms. Therefore, by adding samples sensitive to small perturbations, \dfal enforces the network to increase its ensemble robustness and generalization abilities.

In order to regularize the network and increase its robustness in \dfal, we add both the less robust unlabeled samples and their adversarial attacks. Thus, it is more likely that the network will regularize on the adversarial examples added to the training set and become less sensitive to small adversarial perturbations.
Unlike CEAL, \dfal is hyperparameter-free and cannot corrupt the training set: from the basic definition of adversarial attacks, we know that a sample and its adversarial attack
should share the same label.

Finally \dfal improves the robustness of the network by adding at each iteration unlabeled samples at half the cost of reading their true labels (one label amounts to two samples) as described in Algorithm~\ref{alg:adv_active}.

 \begin{algorithm}
 \caption{\textbf{\textit{DFAL}}: DeepFool Active Learning}
 \label{alg:adv_active}
 \begin{algorithmic}
 \REQUIRE $\mathcal{L}$ set of initial labeled training examples
 \REQUIRE $\mathcal{U}$ set of initial unlabeled training examples
 \REQUIRE $\mathcal{H}$ set of hyper-parameters to train the network
 \REQUIRE $K$ the number of candidates
 \REQUIRE $n_{query}$ the number of data to query
 \REQUIRE p: the $L_p$ norm used ($p=2$)
 \REQUIRE N: the number of data to label\\
 \# init the training set\\
 $ k=0$\\
 $\mathcal{L}_0$ = $\mathcal{L}$\\
 $\mathcal{U}_0$ = $\mathcal{U}$\\
 \WHILE{ k$<$N}
 \STATE \# Train the network $\mathcal{A}_{k}$ given the current labeled training set\\
 $\mathcal{A}_{k}$=training($\mathcal{H},\mathcal{L}_{k})$\\
 \# Select randomly a pool of data $\mathcal{S}_{k}$ of size K\\
 $\mathcal{S}_{k} \subseteq \mathcal{U}_{k};\;\;\mid \mathcal{S}_{k} \mid = K$\\
 \FOR{$x_i \in \mathcal{S}_k$}
 \STATE \#compute adversarial attacks with $L_p$ norms\\
 \STATE $r_i \gets DeepFool(x_i, \mathcal{A}_k;p)$\\
 \ENDFOR\\
 \# query the labels of the $n_{query}$-th samples $\mathcal{Q}_k$ owing the smallest $L_p$ norm perturbation\\
  $index_k\gets argsort(<r_i, r_i>_p \mid i=1..K)$\\
  $\mathcal{Q}_k \gets \{ x_j \mid j\in index_k[0:n_{query}] \} \cup \{ x_j+r_j \mid j\in index_k[0:n_{query}] \}$\\
  $\mathcal{L}_{k+1}\gets \mathcal{L}_k \cup \mathcal{Q}_k$ \\
  $\mathcal{U}_{k+1}\gets \mathcal{U}_k \setminus \mathcal{Q}_k$ \\
  \ENDWHILE
\end{algorithmic}
 \end{algorithm}

\section{Experiments}

\subsection{Dataset and CNN}

We tested our algorithms for fully supervised image classification on three datasets that have been considered in recent articles on active learning for Deep Learning~\cite{huijser2017active}: \mnist, \shoes, and \quick:\\
\textbullet~\mnist: 28x28 grayscale images from 10 digits classes. The training and test set contains respectively 60,000 and 10,000 samples.\\
\textbullet~\shoes: This dataset has been created in \cite{huijser2017active} from the Handbags and the Shoes datasets. It contains RGB images of size 64x64: 184,792 for training
along with 4,000 images for testing.\\
\textbullet~\quick: 28x28 grayscale images from the Google Doodle dataset. We downloaded four classes: Cat, Face, Angel, and Dolphin. 
This lead us to a training set of 444,971 samples and a test set of size 111,246 samples.\\

We assess the efficiency of our method on two CNNs: \lenet and \vgg (\textit{Adam, lr=0.001, batch=32}). We have used Keras and Theano \cite{chollet2015keras, al2016theano}. 
Although we have only tested our methods for CNNs trained with cross-entropy, \dfal may be used on any architectures impaired by adversarial attacks.

\subsection{Evaluation}

We compare the evolution of the test accuracy when querying data with \dfal against the following baselines:

\begin{enumerate}
\item \bald: we select on a random subset of the unlabeled training set, the first $n_{query}$ samples which are expected to maximize the mutual information with the model
parameters. In that order, we sample 10 networks from the approximate posterior of the weights by also applying dropout as test time.\\
 \item \ceal: we select on the whole unlabeled training set, the first $n_{query}$ samples with the highest entropy on their network's prediction. We also
 label any unlabeled samples whose entropy is lower than a given threshold (which is set according to the authors' guidelines: 0.05 for \mnist, 0.19 for \shoes and 0.08 for \quick). Their labels are not queried but estimated from the network's predictions.\\
 \item \core: we select on a random subset of the unlabeled training set, the $n_{query}$ samples which cover at best the training set (labeled and unlabeled data) based on the euclidean distance 
 on the output of the last fully connected layer. To approximate the cover set problem, we follow the instructions prescribed in \cite{sener2018active}: we initialize the selection with the greedy algorithm, and iterate with their Mixed Integer Programming subroutine. We also handle the robustness as prescribed by the authors. We have used \textit{or-tools} 
 \footnote{https://developers.google.com/optimization} to reproduce the MIP subroutine.\\
 \item \egl: we select from a random subset of the unlabeled training set, the first $n_{query}$ samples whose gradients achieves the highest euclidean norm.\\
 \item \uncertainty: we select from the whole unlabeled training set, the first $n_{query}$ samples with the highest entropy on their network's prediction.\\
 \item \random: we select randomly from the whole unlabeled training set $n_{query}$ samples. 
\end{enumerate}

\begin{figure}[!h]
\subfigure[\small{\mnist (\lenet)}]{
   \includegraphics[width=0.225\textwidth, height=0.225\textheight]{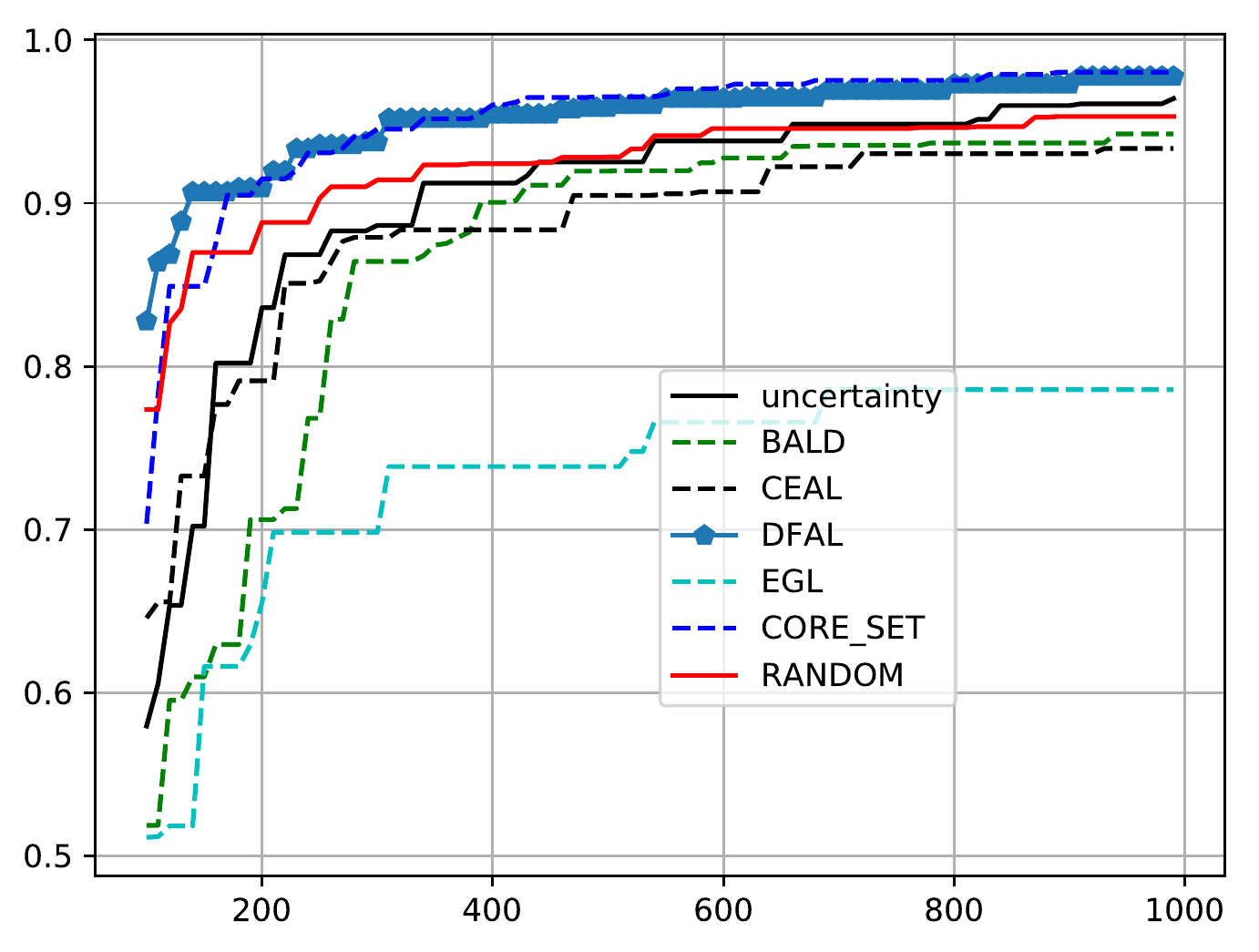}
 \label{sub:mnist_lenet5}}
\subfigure[\small{\mnist (\vgg)}]{
   \includegraphics[width=0.225\textwidth, height=0.225\textheight]{test_acc_MNIST_LeNet5.pdf}
 \label{sub:mnist_vgg8}}
 \subfigure[\small{\shoes (\lenet)}]{
   \includegraphics[width=0.225\textwidth, height=0.225\textheight]{test_acc_MNIST_LeNet5.pdf}
 \label{sub:shoe_bag_lenet5}}
\subfigure[\small{\shoes (\vgg)}]{
   \includegraphics[width=0.225\textwidth, height=0.225\textheight]{test_acc_MNIST_LeNet5.pdf}
 \label{sub:shoe_bag_vgg8}}
 \subfigure[\small{\quick (\lenet)}]{
   \includegraphics[width=0.225\textwidth, height=0.225\textheight]{test_acc_MNIST_LeNet5.pdf}
 \label{sub:quick_draw_lenet5}}
\subfigure[\small{\quick (\vgg)}]{
   \includegraphics[width=0.225\textwidth, height=0.225\textheight]{test_acc_MNIST_LeNet5.pdf}
 \label{sub:quick_draw_vgg8}}
\caption{ Evolution of the test accuracy achieved by 7 active learning techniques( \dfal, \bald, \ceal, \egl, \uncertainty and \random)
given the number of annotations}
\label{fig:convergence}
\end{figure}

We average our results over five trials and plot the accuracy on the test set in figure~\ref{fig:convergence}. Also, we index in table~\ref{tab:convergence} the test accuracy achieved by each active learning methods for fixed size training set: with 100, 500, 800, and 1000 labeled samples.

First of all, an interesting observation is that, independently from networks or datasets, active learning methods originally designed for singleton query 
(\bald, \ceal, \egl, \uncertainty) fail to always compete against random selection (fig~\ref{fig:convergence}). 
This may result from the correlations among the queries when using top score selection. When it comes to our method, \dfal tends to convergence faster than such methods and is always better than random selection, 
independently from the network or the dataset (table~\ref{tab:convergence}). Hence our method is more robust to the hyperparameters settings than other active learning methods, when considering top score selection.

On diverse configurations (\shoes with \lenet and \quick with \vgg), \ceal is worse than uncertainty selection, hence it selects samples with high entropy but mistaken predictions
which adds noise into the training set. Unlike \ceal whose probability of acquiring extra samples depends on the efficiency of the network, \dfal holds a constant number of extra queries, depending only on the number of queries.
Moreover \dfal creates artificial data which are not part of the pool of data. 
For example, in tables~\ref{tab:nb_data_mnist_LeNet5} and ~\ref{tab:nb_data_shoe_bag_LeNet5}, \ceal used more than 20 \% of the training set of \mnist and \shoes, while \dfal only used at most 2 \%.
Thus, \dfal allows more queries, and may also be combined with \ceal.

\begin{table}
\centering
\vspace{-0.5cm}
 \subfigure[\small{\mnist (\lenet)}]{
 \scalebox{0.7}{
 \begin{tabular}{|l|c|c|c|c|c|}
  \hline
  &\multicolumn{5}{c|}{Accuracy (\%)}\\
   \hline
   \# annotations&100&500&800&1000&All\\
   \hline
   \dfal&\textbf{82.77}&96.23&\textbf{97.7}1&98.02&--\\
   \hline
   \bald&51.88&91.96&93.69&94.24&--\\
   \hline
   \ceal&71.81&94.81&96.77&97.33&--\\
   \hline
   \core&78.86&\textbf{96.52}&97.53&\textbf{98.03}&--\\
   \hline
   \egl&58.44&73.86&78.57&78.57&--\\
   \hline
   \uncertainty&57.96&92.52&94.84&96.41&--\\
   \hline
   \random&77.56&92.83&94.63&95.31&\textbf{99.04}\\
   \hline
 \end{tabular}}
 }\vspace{-0.5cm}
  \subfigure[\small{\mnist (\vgg)}]{
  \scalebox{0.7}{
 \begin{tabular}{|l|c|c|c|c|c|}
  \hline
  &\multicolumn{5}{c|}{Accuracy (\%)}\\
   \hline
   \# annotations&100&500&800&1000&All\\
   \hline
   \dfal&\textbf{84.28}&\textbf{96.90}&\textbf{97.98}&\textbf{98.59}&--\\
   \hline
   \bald&53.73&91.47&94.32&94.32&--\\
   \hline
   \ceal&50.87&90.69&90.69&90.69&--\\
   \hline
   \core&78.80&96.68&97.46&97.88&--\\
   \hline
   \egl&37.92&91.84&93.99&93.99&--\\
   \hline
   \uncertainty&45.57&88.36&94.27&94.60&--\\
   \hline
   \random&69.79&91.96&94.05&94.46&\textbf{98.98}\\
   \hline
 \end{tabular}}
 }\vspace{-0.5cm}
  \subfigure[\small{\shoes (\lenet)}]{
 \scalebox{0.7}{
 \begin{tabular}{|l|c|c|c|c|c|}
  \hline
  &\multicolumn{5}{c|}{Accuracy (\%)}\\
   \hline
   \# annotations&100&500&800&1000&All\\
   \hline
   \dfal&94.62&98.50&98.98&99.10&--\\
   \hline
   \bald&93.10&97.95&97.95&97.95&--\\
   \hline
   \ceal&84.65&98.50&99.00&99.12&--\\
   \hline
   \core&92.50&\textbf{98.75}&\textbf{99.07}&\textbf{99.25}&--\\
   \hline
   \egl&75.07&95.47&95.47&95.47&--\\
   \hline
   \uncertainty&\textbf{95.78}&98.35&98.85&98.98&--\\
   \hline
   \random&95.50&98.07&98.07&98.07&\textbf{99.70}\\
   \hline
 \end{tabular}\label{tab:nb_data_shoe_bag_LeNet5}}
 }\vspace{-0.5cm}
  \subfigure[\small{\shoes (\vgg)}]{
  \scalebox{0.7}{
 \begin{tabular}{|l|c|c|c|c|c|}
  \hline
  &\multicolumn{5}{c|}{Accuracy (\%)}\\
   \hline
   \# annotations&100&500&800&1000&All\\
   \hline
   \dfal&\textbf{87.73}&98.53&\textbf{99.30}&99.50&--\\
   \hline
   \bald&86.78&95.35&97.83&97.83&--\\
   \hline
   \ceal&84.20&98.78&99.25&\textbf{99.52}&--\\
   \hline
   \core&0.50&\textbf{99.12}&99.12&99.12&--\\
   \hline
   \egl&0.50&97.28&97.28&97.28&--\\
   \hline
   \uncertainty&83.75&83.75&83.75&83.75&--\\
   \hline
   \random&86.78&95.83&97.08&97.08&\textbf{99.50}\\
   \hline
 \end{tabular}}
 }\vspace{-0.5cm}
 \subfigure[\small{\quick (\lenet)}]{
 \scalebox{0.7}{
 \begin{tabular}{|l|c|c|c|c|c|}
  \hline
  &\multicolumn{5}{c|}{Accuracy (\%)}\\
   \hline
   \# annotations&100&500&800&1000&All\\
   \hline
   \dfal&\textbf{82.56}&\textbf{89.63}&\textbf{90.72}&\textbf{91.09}&--\\
   \hline
   \bald&72.65&87.18&88.34&88.45&--\\
   \hline
   \ceal&70.46&87.04&88.31&89.39&--\\
   \hline
   \core&79.58&88.93&90.54&90.53&--\\
   \hline
   \egl&57.48&64.05&64.05&69.85&--\\
   \hline
   \uncertainty&69.24&86.89&88.54&89.09&--\\
   \hline
   \random&78.09&87.03&88.98&89.42&\textbf{95.46}\\
   \hline
 \end{tabular}}
 }\vspace{-0.5cm}
   \subfigure[\small{\quick (\vgg)}]{
  \scalebox{0.7}{
 \begin{tabular}{|l|c|c|c|c|c|}
  \hline
  &\multicolumn{5}{c|}{Accuracy (\%)}\\
   \hline
   \# annotations&100&500&800&1000&All\\
   \hline
   \dfal&\textbf{84.23}&\textbf{91.52}&\textbf{93.16}&\textbf{93.91}&--\\
   \hline
   \bald&82.00&89.94&91.92&92.87&--\\
   \hline
   \ceal&64.45&79.66&85.73&88.65&--\\
   \hline
   \core&66.71&89.93&92.28&92.62&--\\
   \hline
   \egl&63.12&86.80&90.06&90.06&--\\
   \hline
   \uncertainty&52.77&88.05&89.31&91.03&--\\
   \hline
   \random&78.28&88.13&89.71&89.94&\textbf{96.75}\\
   \hline
 \end{tabular}}
 }
 
%
%
\caption{Test accuracy achieved by 7 active learning techniques for different number of annotations on \lenet and \vgg.}
\label{tab:convergence}
\end{table}

We observe that \dfal always remains in the top three of the best performing active learning methods.
We define those methods based on the test error rate when the labeled training set reaches 1000 samples. When \dfal is outperformed, it is only by a really slight percentage, either by pseudo labeling method( \textit{which contributes more to the training set}),
or by \core. Since \core is designed as a batch active learning strategy, it diminishes the correlations among the queries.
In order to outperform \core, \dfal could be extended into a batch setting approach: instead of selecting the top score samples, one could increase the diversity using for example submodular heuristics \cite{wei2015submodularity}.

Finally, table~\ref{tab:nb_data} compares the effective number of annotations and real number of data required by active learning to reach 
the same test accuracy than when training on the full labeled training set. We only compare \dfal with the best two active learning methods on 1000 samples.
Regarding top score approaches, we notice that \dfal always converges with the smallest number of annotations, on \mnist and \quick. 
When it comes to \shoes, \dfal remains competitive with the core-set approach and \ceal, overall less than 1\% of the training set is needed.

 \begin{table}
  \centering{
  \scalebox{0.6}{\subfigure[\mnist (\lenet)]{
  \begin{tabular}{|c|c|c|}
  \hline
  &\multicolumn{2}{c|}{Accuracy $\geq$ 99.04 \%}\\
  \hline
  &\# annotations&\# labeled data\\
  \hline
  \dfal&\textbf{1210}&2410\\
  \hline
  \small{\core}&1810&\textbf{1810}\\
  \hline
  \small{\ceal}&$\geq$6000&$\geq$6150\\
  \hline
  \end{tabular} \label{tab:nb_data_mnist_LeNet5}
  }}
   \scalebox{0.6}{\subfigure[\mnist (\vgg)]{
  \begin{tabular}{|c|c|c|}
  \hline
  &\multicolumn{2}{c|}{Accuracy $\geq$ 98.98 \%}\\
  \hline
  &\# annotations&\# labeled data\\
  \hline
  \dfal&\textbf{980}&1950\\
  \hline
  \small{\core}&1270&\textbf{1270}\\
  \hline
  \small{\uncertainty}&2800&2800\\
  \hline
  \end{tabular} \label{tab:nb_data_mnist_VGG8}
  }}
  
   \scalebox{0.6}{\subfigure[\shoes (\lenet)]{
  \begin{tabular}{|c|c|c|}
  \hline
  &\multicolumn{2}{c|}{Accuracy $\geq$ 99.70 \%}\\
  \hline
  &\# annotations&\# labeled data\\
  \hline
  \dfal&1070&2130\\
  \hline
  \small{\core}&\textbf{860}&\textbf{860}\\
  \hline
  \ceal&1130&19157\\
  \hline
  \end{tabular}
  }}
  \scalebox{0.6}{\subfigure[\shoes (\vgg)]{
 \begin{tabular}{|c|c|c|}
 \hline
 &\multicolumn{2}{c|}{Accuracy $\geq$ 99.50 \%}\\
 \hline
 &\# annotations&\# labeled data\\
 \hline
 \dfal&530&1050\\
 \hline
 \small{\core}&\textbf{400}&\textbf{400}\\
 \hline
 \ceal&580&705\\
 \hline
 \end{tabular}
 }}
 
 \scalebox{0.6}{\subfigure[\quick (\lenet)]{
 \begin{tabular}{|c|c|c|}
 \hline
 &\multicolumn{2}{c|}{Accuracy $\geq$ 95.46\%}\\
 \hline
 &\# annotations&\# labeled data\\
 \hline
 \dfal&\textbf{7470}&14930\\
 \hline
 \small{\core}&$\geq$8590&$\geq$8590\\
 \hline
 \small{\uncertainty}&$\geq$10590&$\geq$10590\\
 \hline
 \end{tabular}
 }}
  \scalebox{0.6}{\subfigure[\quick (\vgg)]{
 \begin{tabular}{|c|c|c|}
 \hline
 &\multicolumn{2}{c|}{Accuracy $\geq$  96.75\%}\\
 \hline
 &\# annotations&\# labeled data\\
 \hline
 \dfal&\textbf{4810}&9610\\
 \hline
 \small{\core}&$\geq$6750&$\geq$6750\\
 \hline
 \bald&5590&5590\\
 \hline
 \end{tabular}
 }}
  \caption{Comparison of the number of annotations and effective data required to achieve the same test accuracy on \lenet and \vgg as the accuracy obtained on the full training set ($\pm$ 0.5). 
  We considered \dfal against the active methods achieving best accuracy on 1000 samples.}
   \label{tab:nb_data}
  }
 \end{table}

\subsection{Comparative study between \dfal and the \core approach}

In most of our experiments, \dfal is competitive with the current state-of-the-art method, \core, sometimes outperforming it by a large margin (tab~\ref{tab:nb_data_mnist_LeNet5},\ref{tab:nb_data_mnist_VGG8}).
On the other hand, our method is more interesting than \core when considering the computational time. 
Indeed one of the main cons raised against \core is that the optimal solution is a NP-Hard problem. 
To overcome this issue, the authors used a greedy solution, which is known to hold a 2-OPT bound. 
Then, they optimize this solution, using a Mixed Integer Programming subroutine on which they iterate to improve the coverage. 
While constructing this MIP, they also handle the weakness of k-center, namely robustness: they assume an upper limit on the number of outliers.
However, using robustness, as prescribed in the original paper, slows down the active selection.
Their solution selects a batch of data at each time, while our method attributes scores to each unlabeled sample independently one from another. 
Hence \dfal can be easily parallelized to compute adversarial attacks for a large pool of unlabeled samples.

We demonstrate the computational time gap between our method, \dfal, and \core in table~\ref{tab:complexity}: we have recorded the average runtime of selecting 10 queries on \mnist with a training set of 100 samples and an unlabeled pool of size 800. 
For a sake of fairness, we compare \dfal running time against the \core approach, 
with and without robustness \footnote{Intel(R) Xeon(R) CPU E5-2670 v3 @ 2.30GHz; 64 GB memory and GTX TITAN X}. Notice that the runtime performance of \dfal is independent from the size
of the labeled training set. While \core slows down while we add more and more data to the training set.

\begin{table}
\centering
\scalebox{0.8}{\begin{tabular}{|c|c|c|c|}
\hline
 &\dfal& \core& \core\\
 &&(with regularisation)&(no regularisation)\\
 \hline
 second&\textbf{126.54}&891.78&784.99\\
 \hline
\end{tabular}}
\caption{Average runtime of \dfal and \core on \mnist: $\mid \mathcal{L} \mid=100$, $\mid \mathcal{U} \mid=800$, $n_{query}=10$}
 \label{tab:complexity}
\end{table}

\subsection{Transferability}

\begin{figure}[!h]
\includegraphics[width=0.45\textwidth, height=0.25\textheight]{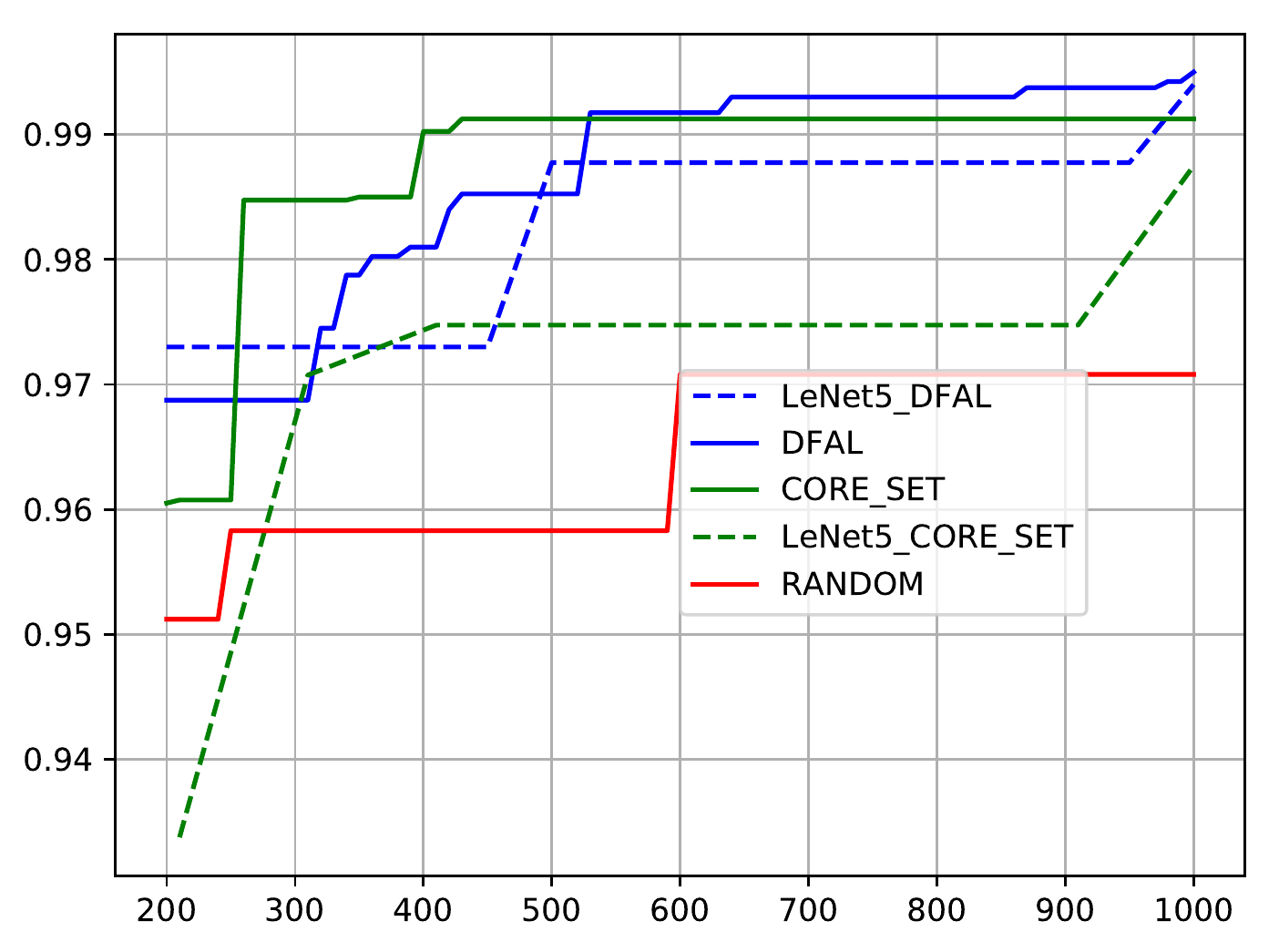}
\caption{Evolution of the test accuracy for (\shoes, \vgg) trained with different labeled training set: we compare the efficiency of \dfal and \core built on \lenet and transfered on \vgg.
The data selected by \dfal for \lenet achieve better test accuracy than the data transfered from \core. At 1000 samples, they converge to similar test accuracy than the data specifically designed for \vgg.}
\label{fig:transferability}
\end{figure}

\begin{table}
\scalebox{0.8}{\subfigure[\mnist]{
 \begin{tabular}{|c|c|c|c|}
 \hline
  &\dfal&\core&\random\\
  \hline
  \lenet $\rightarrow$ \vgg&\textbf{97.80}&96.90&94.46\\
  \hline
  \vgg $\rightarrow$ \lenet&\textbf{97.93}&97.40&95.31\\
  \hline
 \end{tabular}}}
 \scalebox{0.8}{\subfigure[\quick]{
 \begin{tabular}{|c|c|c|c|}
 \hline
  &\dfal&\core&\random\\
  \hline
  \lenet $\rightarrow$ \vgg&\textbf{92.87}&91.06&89.94\\
  \hline
  \vgg $\rightarrow$ \lenet&89.23&89.41&\textbf{89.42}\\
  \hline
 \end{tabular}}}
 \scalebox{0.8}{\subfigure[\shoes]{
 \begin{tabular}{|c|c|c|c|}
 \hline
  &\dfal&\core&\random\\
  \hline
  \lenet $\rightarrow$ \vgg&\textbf{99.40}&99.12&97.08\\
  \hline
  \vgg $\rightarrow$ \lenet&\textbf{98.75}&98.50&98.07\\
  \hline
 \end{tabular}}}
 \caption{Comparison of the transferability of \dfal and \core with 1000 annotations}
 \label{tab:test_transferability}
\end{table}

In preliminary experiments to a new problem, we know in advance neither the model architecture nor the hyperparameters that are best suited for the problem.
One can argue that a network with high capacity is likely to give high accuracy and is sufficient enough when combined with some human expertise on the problem: several architectures have been handcrafted for specific tasks and are available online. Still, their efficiency is known under typical training procedures and with large datasets. In \cite{shen2018deep}, Yanyao Shen \etal pointed out an interesting flaw in active learning: they succeed in outperforming classical methods for Named Entity Recognition using only 25\% of the training set but by introducing a lightweight architecture. Hence, when using a single predefined model, active learning may optimize the training set to a model not well optimized for the task at hand. Such an issue is inherent to active learning. Combining model selection with active learning has been investigated for shallow models.
One of the main issue raised is that multiple hypotheses trained in parallel may benefit from labeling different training points. Hence an active learning strategy effective on any fixed model may be less efficient than random sampling when considering it with model selection.
Although combining model selection and active learning for any type of model is non-trivial,
deep learning owns a specific property: the transferability of adversarial examples towards a wide range of architectures lead to assume that the decision borders of neural networks trained on similar tasks overlap.


\dfal overcomes this limitation. Indeed it is well known that adversarial attacks handcrafted for a specific network may be used with success
on other networks, especially when considering CNNs.
The reason raised is that the distance between network's decision borders is smaller than most adversarial perturbations.
Based on that argument, we may assume that most of the \dfal queries are useful for a diverse set of architectures, not only the one they have been queried for.

When it comes to the transferability, we empirically demonstrate \dfal's potential on a baby task: in figure~\ref{fig:transferability}
we recorded \shoes adversarial queries for \lenet and use
them for training \vgg. While the test accuracy achieved is lower than with the adversarial active queries designed for \vgg, the transfered training set achieves better accuracy than random selection, but also, when reaching 1000 annotated samples, it is also better than queries from other active criteria designed for \vgg.
We go further and compare the test accuracy of \dfal and \core transfered dataset on 1000 samples in table~\ref{tab:test_transferability}. 
Surprisingly the transfered queries from \core perform better than random. However, in almost every case,
the transfered queries from \dfal outperform \core and \random. 
The only exception concerns the transfered queries from \vgg to \lenet: neither \dfal nor \core succeed in outperforming \random. We believe that \lenet trained on \quick have a smoother decision boundary than \vgg in our hyperparameter setting. 
Thus, it would result in \vgg queries being useful for training \lenet, while the opposite would not be true.

\section{Conclusion}
%
%
In this paper, we propose a new heuristic, \dfal, to perform margin based active learning for CNNs: we approximate the projection of a sample to the decision boundary by its smallest adversarial
attack.
We demonstrate empirically that our adversarial active learning strategy is highly efficient for CNNs trained on \mnist, \shoes, and \quick.
Not only we are competitive with the state-of-the-art batch active learning method for CNNs, \core, but we also outperform \core for runtime performance.
Thanks to the transferability of adversarial attacks, \dfal is a promising approach for combining active learning with model selection for deep networks

\bibliographystyle{icml2018}
\bibliography{biblio}

\end{document}